# From Phenomenological Fitting to Endogenous Deduction: A Paradigm Leap via Meta-Principle Physics Architecture

Author: Helong Hu[1], HongDan Pan[2], ShuiQing Hu[2]

Affiliation:

[1] Guangdong Ocean University, No.1 Luqin Road, Yangjiang City, Guangdong Province, China（524088）

Email: 604232736@qq.com

[2] Maoming Administrative Service Center, Compound 6, 10th Youcheng Road, Maonan District, Maoming City, Guangdong Province, China (525200)

Email: 1226997226@qq.com

[2] China Pacific Insurance (Group) Co., Ltd. Foshan Branch, No.17, Jihua 5th Road, Chancheng District, Foshan City, Guangdong Province, China (528000)

Email: 278990760@qq.com

## Abstract

The essence of current neural network architectures is phenomenological fitting: they learn the statistical correlations between inputs and outputs through massive parameters and data, yet lack an intrinsic understanding of the fundamental principles governing physical reality. This paper proposes a paradigm leap from pure phenomenological fitting to the fusion of phenomenological fitting and endogenous deduction. By explicitly embedding the meta-principles of the physical world into the neural network architecture, we construct the Meta-Principle Physics Architecture (MPPA).

Specifically, MPPA explicitly embeds three core meta-principles of the physical world—the Principle of Connectivity, the Principle of Conservation, and the Principle of Periodicity—into the neural network architecture, implemented through three core components: the Gravitator realizes the Principle of Connectivity based on a standard causal attention mechanism; the Energy Encoder implements the Principle of Conservation via log-domain energy tracking and delayed compensation mechanism; the Periodicity Encoder fulfills the Principle of Periodicity through FFT-based spectral analysis and delayed modulation mechanism. These components work collaboratively via a learnable independent gating fusion mechanism, forming a complete physical cognition framework of "local relational connectivity - global conservation constraint - evolutionary periodic law".

Experimental results show that MPPA achieves a significant improvement in physical reasoning capability (from nearly zero to 0.436, 0.436 vs 0.000), a 2.18-fold improvement in mathematical tasks (0.330 vs 0.151), a 52% improvement in logical tasks (0.456 vs 0.300), and a 3.69% reduction in validation perplexity (259.45 vs 269.40), with only an 11.8% increase in parameter count (242.40M vs 216.91M). More importantly, MPPA exhibits strong generalization ability on out-of-distribution physical scenarios, demonstrating the robustness and interpretability of the principle-embedded architecture. This study provides a new theoretical foundation and technical path for building the next generation of artificial intelligence systems with physical common sense, causal reasoning ability, and mathematical rigor.

# 1. Introduction

Over the past decade, the field of artificial intelligence has achieved milestone advances. Large-scale autoregressive neural network architectures represented by Transformer[1] have demonstrated unprecedented performance on general tasks such as natural language understanding, computer vision, and code generation. BERT[2] ushered in a new era of pre-trained language models through a bidirectional context encoding architecture, while the GPT series[3][4][11] systematically verified the general capability of generative pre-training in zero-shot and few-shot settings. However, such models have always had an essential limitation: their core paradigm belongs to phenomenological fitting, which learns statistical correlations between inputs and outputs through massive parameters and large-scale training data, and cannot endogenously model the causal mechanisms and underlying physical principles behind the phenomena[5][6].

This limitation is fully exposed when facing tasks that require rigorous physical reasoning, mathematical proof, or counterfactual inference. For example, when generating a description of planetary motion, traditional language models perform probabilistic combination based on similar texts seen in the training corpus, rather than deductive reasoning based on an intrinsic understanding of the law of universal gravitation and the principle of energy conservation. As pointed out by Bender et al.[6], purely statistical learning-based models have a fundamental "stochastic parrots" risk: they can generate fluent text, yet lack a true understanding of semantics and grasp of causal relationships.

## 1.1 Paradigm Shift from Phenomenological Fitting to Endogenous Deduction

The core mission of physics is to uncover the universal laws of nature. From Newton's law of universal gravitation to Einstein's general relativity, from Maxwell's equations to the Schrödinger equation of quantum mechanics, every major breakthrough in physics stems from a profound insight into "meta-principles". These meta-principles—such as causal connectivity, energy conservation, and periodic symmetry—form the cornerstone of understanding the physical world. They are not only mathematical formulas describing phenomena, but also causal mechanisms with universal explanatory power, which can support counterfactual reasoning and out-of-distribution generalization[7][8].

The core argument of this paper is that to realize the paradigm leap of artificial intelligence from phenomenological fitting to endogenous deduction, the scientifically validated meta-principles of the physical world must be explicitly and structurally embedded into the fundamental architecture of neural networks. Accordingly, we propose the Meta-Principle Physics Architecture (MPPA), which strictly maps the three core meta-principles of the physical world—the Principle of Connectivity, the Principle of Conservation, and the Principle of Periodicity—into differentiable and learnable modular components of neural networks. While retaining the general fitting capability of pre-trained models, it endows the model with endogenous deductive ability of physical laws.

## 1.2 Philosophical Foundation and Computational Implementation of Meta-Principles

Our selection of connectivity, conservation, and periodicity as the three core meta-principles is not a subjective setting, but based on an in-depth of the ontological structure of the physical world and human cognitive logic. The three form a logically self-consistent, hierarchically complementary complete cognitive system:

(1) **Principle of Connectivity** — From a local perspective, there are interactions between things. No object exists in isolation; it establishes connections with other objects through gravity, electromagnetic force, contact force, and other interactions. The success of Transformer[1] precisely lies in capturing this connectivity: the self-attention mechanism is essentially a soft-weighted connectivity modeling, enabling each token to adaptively capture dependencies with other tokens. However, connectivity alone is insufficient to understand the physical world—it only answers "how things interact", but cannot explain the two core questions of "how the system as a whole remains stable" and "what laws the system evolution follows".

(2) **Principle of Conservation** — From a holistic perspective, a closed system has the property of keeping certain total quantities constant. Energy conservation, momentum conservation, charge conservation, and others are the cornerstones of physics. Conservation provides a global constraint for connectivity: despite continuous local interactions, the total quantity of the system must remain balanced. This tension between "local dynamics and global statics" is a key feature of the physical world. As Noether[9] proved in her groundbreaking theorem: *"Jede kontinuierliche Symmetrie der Wirkung entspricht einer Erhaltungsgröße."* (For every continuous symmetry of the action, there corresponds a conserved quantity.)

(3) **Principle of Periodicity** — The combined action of connectivity and conservation produces regularity. When various parts of the system are interconnected and constrained by conservation laws, the evolution of the system inevitably presents a regular (periodic) evolution pattern. The periodic orbit of a planet is the result of the combined action of gravity (connectivity) and angular momentum conservation (conservation); the simple harmonic motion of a spring oscillator is the product of elastic force and energy conservation. As Feynman[10] pointed out in his famous lecture notes: *"The same equations have the same solutions."* Periodicity is precisely the external manifestation of the internal structure of physical equations.

The logical relationship of the meta-principles constitutes a logically self-consistent philosophical system, forming a complete physical cognition framework of "local relational connectivity - global conservation constraint - evolutionary periodic law".

### 1.3 Main Contributions

The main contributions of this paper include:

(1) **Theoretical Contributions**

We propose a new paradigm of "endogenous deduction" for artificial intelligence, establish a complete theoretical framework from pure data-driven phenomenological fitting to the deep fusion of phenomenological fitting and endogenous deduction, and clarify the core path for the evolution of artificial intelligence from "statistical memory" to "principle understanding".

We establish the core architectural positioning of MPPA as "dual-wheel driven by both phenomenological fitting capability and endogenous deduction capability", and clarify the synergistic relationship between general language modeling and professional physical deduction.

We establish a strict one-to-one mapping between the philosophical foundation, physical connotation, and differentiable computational implementation in neural networks of the three physical meta-principles (connectivity - conservation - periodicity), providing a standardized theoretical paradigm for embedding physical principles into neural network architectures.

(2) **Architectural Innovations**

We design a Chunk-based Mechanism under causal constraints, which realizes efficient block-wise parallel processing of long sequences and cross-block information transfer under strict causal constraints, resolving the core contradiction between global physical principle embedding, causal constraints of autoregressive models, and parallel training efficiency.

We design a Causal Delayed Compensation mechanism, which realizes the endogenous embedding of global physical constraints while strictly following the autoregressive causal order, filling the technical gap that existing physics-informed neural networks cannot adapt to autoregressive generation scenarios.

We propose two new modular neural network components with clear physical connotations, the Energy Encoder and the Periodicity Encoder, which respectively realize the differentiable computation of the conservation principle and the periodicity principle, providing a reusable modular solution for embedding physical principles into neural network architectures.

## 2. Related Work

### 2.1 Pre-trained Language Models

The development of pre-trained language models has completely transformed the field of natural language processing. BERT[2] learns deep bidirectional context representations through masked language modeling and next sentence prediction tasks, achieving breakthrough progress in multiple NLP tasks. GPT[3] adopts autoregressive generative pre-training, demonstrating the potential of large-scale language models in zero-shot learning. GPT-2[4] further scales to 1.5 billion parameters, proving the positive correlation between model scale and performance. GPT-3[11] increases the parameter scale to 175 billion, exhibiting amazing few-shot learning ability. RoBERTa[12] further improves the pre-training effect by optimizing the training strategy of BERT. T5[13] unifies all NLP tasks into a text-to-text conversion framework, simplifying the design of model architecture.

However, these models are still essentially phenomenological fitting systems, lacking an intrinsic understanding of the basic principles of the physical world.

### 2.2 Physics-Informed Neural Networks

Physics-Informed Neural Networks (PINNs) were formally proposed by Raissi et al.[14] in 2019. Their core idea is to embed known physical governing equations (such as the Navier-Stokes equations, heat conduction equations) into the loss function as soft constraints, enabling the neural network to fit the observed data while satisfying the physical laws. PINNs have achieved remarkable success in fluid mechanics, materials science, geophysics and other fields[15], but their limitations lie in: the specific form of the physical equation needs to be known in advance, it is difficult to deal with complex multi-physics coupling problems, and the generalization ability is limited by the distribution of training data.

Karniadakis et al.[15] provided a comprehensive review of physics-informed machine learning, demonstrating its great potential in scientific computing. However, the essence of PINNs is still phenomenological fitting plus external constraints: physical knowledge acts on the loss level, rather than being embedded into the network architecture itself. As critics have pointed out, the internal representation of the model remains arbitrary, only being "pulled" towards the direction that satisfies the physical constraints by the loss function.

The SINDy (Sparse Identification of Nonlinear Dynamics) method[16] sparsely identifies the governing equations of nonlinear dynamic systems from observed data, but it still belongs to the category of "post-hoc added constraints" and does not integrate physical principles into the basic computational flow of the model. The fundamental limitation of the above methods is that physical principles are not the core component of the model's endogenous computational mechanism, but external constraints added post-hoc during the training process.

The essential differences between MPPA and the above methods can be summarized into three dimensions:

**Different constraint levels**: PINN-like methods only impose physical constraints on the model output at the loss function level, while MPPA directly embeds physical meta-principles into the fundamental architecture of the neural network, realizing endogenous modeling of physical principles.

**Different causal processing**: PINN-like methods usually perform global optimization based on complete time trajectories and cannot meet the strict causal constraints of autoregressive modeling; all components of MPPA are designed under strict causal constraints and can be directly adapted to autoregressive generation scenarios.

**Different paradigm fusion**: MPPA realizes the deep fusion of phenomenological fitting capability and endogenous deduction capability. While retaining the general language modeling capability of pre-trained models, it obtains endogenous physical deduction ability. In contrast, PINN-like methods are usually designed only for specific physical scenarios and lack adaptability to general tasks.

### 2.3 Neural Ordinary Differential Equations

Neural Ordinary Differential Equations (Neural ODEs) were proposed by Chen et al.[17]. Their core innovation is to model the forward propagation process of neural networks as a continuous-time ordinary differential equation solving process, achieving constant memory overhead and continuous latent space evolution modeling through a reversible ODE solver. Neural ODEs provide an elegant continuous-time mathematical framework for learning dynamic systems, but they are mainly oriented to the fitting of continuous-time systems, and have obvious limitations in modeling the causal structure of discrete sequences and endogenous constraints of conservation properties, making it difficult to directly adapt to autoregressive language modeling scenarios.

### 2.4 Causal Inference and Structure Learning

Causal inference aims to identify causal relationships between variables from observed data, rather than just statistical correlations. Pearl[8] proposed causal graph models and do-calculus, laying the mathematical foundation for formal causal reasoning. Schölkopf et al.[18] pointed out that causal representation learning is the key to realizing robust artificial intelligence, emphasizing the importance of the causal structure of the data generation process for generalization ability.

However, most existing causal inference methods focus on "discovering" potential causal structures from observed data, and do not explicitly and structurally embed scientifically validated prior causal principles into the fundamental computational architecture of neural networks. The core innovation and uniqueness of MPPA lie in: taking physical meta-principles verified by hundreds of years of scientific practice as core prior knowledge, directly embedding them into the fundamental architecture of neural networks, so that the model has the correct underlying logic of causal reasoning from the initialization stage, rather than learning causal relationships from data from scratch.

### 2.5 Attention Mechanism and Transformer Architecture

The Transformer architecture was proposed by Vaswani et al.[1], which realizes global dependency modeling between sequence elements through the self-attention mechanism, completely changing the fields of natural language processing and computer vision. The essence of the self-attention mechanism is to calculate the similarity between the Query and Key, and perform weighted aggregation on the Value accordingly. Its mathematical form is:

$$\text{Attention}(Q, K, V) = \text{softmax}\left(\frac{QK^T}{\sqrt{d_k}}\right)V \quad (1)$$

where $d_k$ is the dimension of the key vector. The success of Transformer has proven the strong ability of the attention mechanism in capturing long-distance dependencies, but its essence is still weighted aggregation based on statistical correlation, lacking explicit modeling of causal directionality and physical conservation laws[19].

### 2.6 Efficient Transformer Variants

The self-attention mechanism of the standard Transformer has $O(n^2)$ time and space complexity, which limits its ability to process long sequences. To this end, researchers have proposed various efficient variants. Longformer[20] adopts a strategy combining local window attention and global attention, reducing the complexity to linear. BigBird[21] theoretically proves its Turing completeness through the combination of random attention, window attention and global attention. Reformer[22] uses Locality Sensitive Hashing (LSH) to approximate nearest neighbor search, reducing the attention complexity to $O(n\log n)$. Performer[23] approximates the softmax kernel through Orthogonal Random Features (ORF), realizing attention calculation with linear complexity. Tay et al.[24] conducted a comprehensive review of efficient Transformers, systematically classifying various optimization strategies.

However, the core optimization goal of the above efficient Transformer variants is to reduce computational complexity and improve long sequence processing capability, without involving explicit embedding and endogenous modeling of physical principles. While realizing the endogenous embedding of physical meta-principles, MPPA achieves efficient sequence parallel processing through the chunk-based mechanism, taking into account both physical deduction capability and computational efficiency.

### 2.7 State Space Models

State space models such as Mamba[25] realize sequence modeling with linear time complexity through selective state spaces, providing an efficient solution for long sequence processing. Katharopoulos et al.[26] proved that Transformer can be regarded as a form of RNN, and sequence modeling with linear complexity can be realized through linear attention mechanism. However, similar to efficient Transformer variants, the core optimization goal of such models still focuses on computational efficiency and long sequence modeling capability, without exploring the explicit embedding and endogenous modeling of physical principles.

### 3. Meta-Principle Physics Architecture (MPPA)

### 3.1 Architecture Overview

The core design concept of the Meta-Principle Physics Architecture (MPPA) is to explicitly embed the three meta-principles of the physical world—the Principle of Connectivity, the Principle of Conservation, and the Principle of Periodicity—into the neural network architecture, so that the model has endogenous deductive ability of physical laws, rather than only memorizing statistical patterns in the training data. The overall architecture of MPPA is shown in Figure 1, which includes three core components:

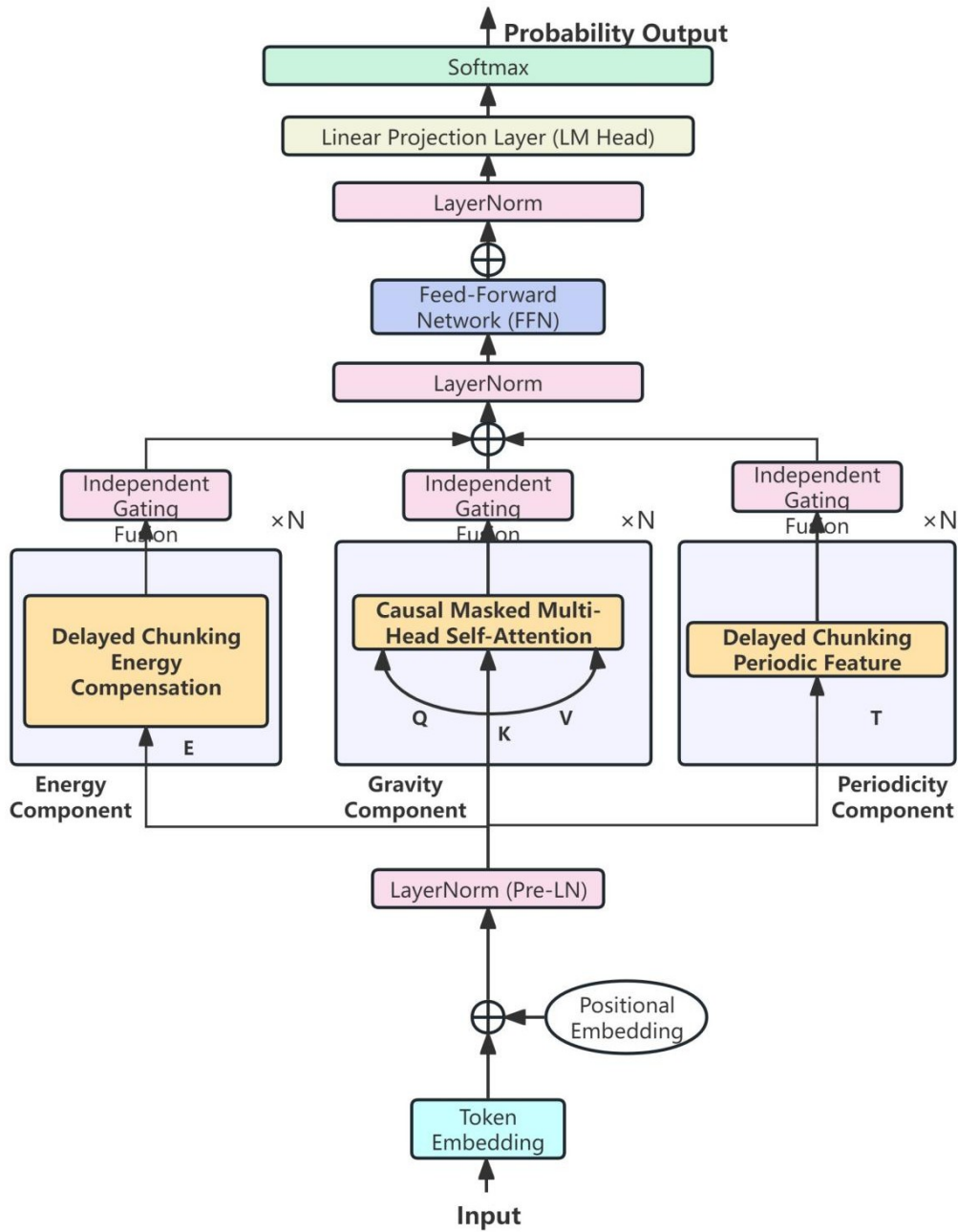

**Figure 1 Meta-Principle Physics Architecture (MPPA) Schematic**

**Gravitator**: Realizes the Principle of Connectivity based on the standard causal attention mechanism, capturing the causal dependencies between sequence elements.

**Energy Encoder**: Implements the Principle of Conservation via log-domain energy tracking and delayed compensation mechanism, ensuring the conservation of information during propagation.

**Periodicity Encoder**: Fulfills the Principle of Periodicity through FFT-based spectral analysis and delayed modulation mechanism, modeling the periodicity and oscillatory behavior of the system.

These three components work collaboratively through an independent gating fusion mechanism. Let the input sequence be $x \in \mathbb{R}^{n \times d}$, where $n$ is the sequence length and $d$ is the word embedding dimension. The outputs of the three meta-principle components are $Z$ (Gravitator), $E$ (Energy Encoder), and $T$ (Periodicity Encoder) respectively, then the final output after module fusion is:

$$\mathbf{H}_{\text{out}} = \mathbf{x} + g_g \cdot \mathbf{Z} + g_e \cdot \mathbf{E} + g_p \cdot \mathbf{T} \quad (2)$$

where $g_g$, $g_e$, $g_p$ are learnable gating parameters, which are used to dynamically control the contribution weight of each meta-principle component. This design enables the model to adaptively adjust the activation intensity of different physical principles according to the input content and the requirements of downstream tasks, taking into account both general modeling capability and professional physical deduction ability.

### 3.2 Gravity Component: Implementation of Connectivity Principle

The Principle of Connectivity is one of the cornerstones of physics, which states that there are universal causal connections between physical entities. From Newton's law of universal gravitation to quantum entanglement, from electromagnetic interaction to strong and weak nuclear forces, all physical processes in nature reflect the interconnection and causal dependence between things.

The Gravitator realizes the Principle of Connectivity through the standard causal attention mechanism. Like the self-attention of the standard Transformer, it adopts causal masking to ensure that each position can only attend to its previous positions, thus modeling strict temporal causality. Given the input $H \in \mathbb{R}^{n \times d}$, the calculation process of the Gravitator is:

$$Q = HW_Q, K = HW_K, V = HW_V \quad (3)$$

$$S = \frac{QK^T}{\sqrt{d_k}} \quad (4)$$

$$\mathbf{M}_{ij} = \begin{cases} 0, & \text{if } i \geq j \\ -\infty, & \text{if } i < j \end{cases} \quad (5)$$

$$\mathbf{A} = \text{softmax}(\mathbf{S} + \mathbf{M}) \quad (6)$$

$$Z = AV \quad (7)$$

where $W_Q$, $W_K$, $W_V$ are learnable linear projection matrices, $d_k$ is the dimension of the key vector, and $M$ is the lower triangular causal mask matrix. This design strictly ensures that when the model predicts the output of the $i$-th position, it can only use the historical information from the 1st to the $(i-1)$-th position, fully complying with the causal order constraint of autoregressive modeling.

### 3.3 Energy Component: Implementation of Conservation Principle

The Principle of Conservation is one of the most profound and universal meta-principles in physics. From energy conservation to momentum conservation, from angular momentum conservation to charge conservation, conservation laws run through all branches of physics. Noether's theorem[9] reveals the profound connection between conservation laws and symmetry: every continuous symmetry corresponds to a conserved quantity.

The Energy Encoder is based on the Energy Conservation Principle, and realizes conservation maintenance under strict causal constraints through chunk-based processing and delayed compensation.

### 3.3.1 Chunk-based Mechanism

The chunk-based mechanism is one of the core innovations of MPPA, which resolves the core contradiction between global physical principle embedding, strict causal constraints of autoregressive models, and parallel training efficiency. This mechanism divides the input sequence into fixed-size chunks, each of which can be processed in full parallel internally, and information transfer between chunks under strict causal constraints is realized through the delayed compensation mechanism. This design not only retains the high efficiency of parallel training of the Transformer architecture, but also realizes cross-chunk global information transfer and conservation constraints. Let the chunk size be $C$, the input sequence $H \in \mathbb{R}^{n \times d}$ is divided into $N = \lceil n/C \rceil$ consecutive chunks:

$$H = [h_1, h_2, ..., h_N] \quad (8)$$

where each chunk $h_i \in \mathbb{R}^{C \times d}$, and the last chunk is padded with zeros to the fixed length $C$ if its length is insufficient.

### 3.3.2 Log-Domain Energy Tracking

The core design idea of the Energy Encoder is: during the propagation of information flow in the neural network, information should not be generated or disappeared without reason, but should maintain a certain form of total conservation, which is directly derived from the law of energy conservation in the physical world. The Energy Encoder tracks the change of the "energy" of the hidden state in each chunk, and performs causal compensation in subsequent chunks to ensure the conservation of the information flow of the entire sequence.

Specifically, the Energy Encoder first calculates the average energy of each chunk:

$$E_i = \frac{1}{C} \sum_{t=(i-1)C+1}^{iC} log(1 + |h_t|^2) \quad (9)$$

where $h_t$ is the hidden state vector of the $t$-th position in the sequence, and $|\cdot|$ is the L2 norm. Subsequently, the Energy Encoder maintains the cumulative energy sum of historical chunks in the log domain:

$$L_i = \sum_{j=1}^{i-1} log(E_j + \varepsilon) \quad (10)$$

where $\varepsilon$ is a very small constant to ensure numerical stability (default value is $10^{-6}$).

### 3.3.3 Delayed 2-Step Compensation Mechanism

The global constraint of energy conservation requires estimating the global mean based on the energy distribution of historical chunks, while the causal constraint requires that future information cannot be used. The delayed compensation mechanism is precisely designed to solve this core contradiction. This mechanism strictly guarantees:

The compensation of the current chunk only uses the calculated and stable historical deviation information;

It fully meets the causal constraints of autoregressive modeling, and absolutely does not use the information of future chunks;

Cross-chunk global state transfer is realized through the historical deviation (debt) queue.

First, the global geometric mean energy is calculated based on the historical cumulative energy:

$$E_i = exp\left(\frac{L_i}{i-1}\right) \tag{11}$$

Subsequently, the "debt" (i.e., the deviation value) of the current chunk energy relative to the global geometric mean is calculated:

$$D_i = E_i - E_i \tag{12}$$

The calculated debt is stored in the historical queue, and is used to compensate subsequent chunks after a 2-step delay. The compensated hidden state of the chunk is:

$$H_e^{(i)} = h_i \cdot exp(\sigma(D_{i-2}) \cdot I) \tag{13}$$

where $\sigma$ is the Sigmoid activation function, and $I$ is the learnable compensation intensity parameter. The 2-step delay design ensures that the compensation uses the deviation of the historical chunks that have been fully calculated, rather than the information of the chunk currently being calculated, thus strictly meeting the causal constraints of autoregressive modeling.

### 3.3.4 Implementation Details and Historical Information Representation

The Energy Encoder expresses historical information in the current output through the following mechanisms:

Log-domain accumulation: maintaining the log sum of the energy of all chunks from the beginning of the sequence to the current

Geometric mean estimation: estimating the global energy level based on the historical energy distribution

Debt queue: storing the deviation between historical chunks and the global mean

Delayed feedback: feeding back historical deviations to the current chunk to achieve energy balance

### 3.4 Periodicity Component: Implementation of Periodicity Principle

The Principle of Periodicity is a universal law describing oscillation and cyclic phenomena in nature. From planetary orbits to pendulum motion, from electromagnetic oscillation to biological rhythms, periodic phenomena are ubiquitous. Fourier analysis[27] reveals that any periodic signal can be decomposed into the superposition of sine waves of different frequencies, providing a mathematical tool for understanding complex periodic phenomena.

Based on the core connotation of the Periodicity Principle, the Periodicity Encoder realizes endogenous modeling of system periodicity under strict causal constraints through chunk-level FFT-based spectral analysis and causal delayed modulation mechanism.

### 3.4.1 Chunk-level FFT Spectral Analysis

The calculation process of the Periodicity Encoder is as follows: first, the Fast Fourier Transform (FFT) is applied to each chunk $h_i \in \mathbb{R}^{C \times d}$ along the sequence dimension:

$$\mathbf{F}_i = \text{FFT}(\mathbf{h}_i, \dim = 0) \tag{14}$$

where $F_i \in \mathbb{C}^{C \times d}$ is the complex spectrum. Then, the magnitude spectrum of the spectrum is calculated:

$$A_i = |F_i| \qquad (15)$$

### 3.4.2 EMA Spectral Accumulation and Delayed 1-Step Modulation

The Periodicity Encoder accumulates historical spectral information through the Exponential Moving Average (EMA) mechanism:

$$S_i = \alpha \cdot S_{i-1} + (1 - \alpha) \cdot A_i \qquad (16)$$

where $\alpha$ is the learnable decay coefficient (constrained between 0.1 and 0.9 through Sigmoid). The EMA mechanism is essentially an autoregressive filter, which accumulates the frequency features of historical chunks into the current state.

The delayed 1-step modulation mechanism uses the EMA state of the previous chunk to modulate the current chunk:

$$M_i = MLP(S_{i-1}) \qquad (17)$$

$$H_p^{(i)} = h_i \odot M_i \qquad (18)$$

where $\odot$ represents element-wise multiplication, and the MLP projects the spectral features into modulation parameters (scale/shift). The 1-step delay design ensures that the modulation uses the stable historical spectral information, strictly satisfying the causal constraint.

### 3.4.3 Implementation Details and Historical Information Representation

The Periodicity Encoder expresses historical information in the current output through the following mechanisms:

EMA spectral accumulation: maintaining the exponential moving average of the spectrum of historical chunks

Learnable decay rate: independently learning the retention ratio of historical information for each dimension

Delayed modulation: using historical spectral information to modulate the current chunk

Frequency domain projection: projecting spectral features into modulation parameters (scale/shift)

The essence of historical information representation: the EMA mechanism of the Periodicity Encoder is essentially an autoregressive filter, which accumulates the frequency features of historical chunks into the current state. This accumulation enables the current output to not only reflect the characteristics of the current chunk, but also carry the frequency evolution information of the entire historical sequence.

### 3.5 Gating Fusion Mechanism

The gating fusion mechanism is one of the core innovations of MPPA, which enables the model to adaptively adjust the contribution degree of the three meta-principle components. Different from simple weighted average, the gating fusion mechanism dynamically controls the activation intensity of each component through learned gating parameters, realizing selective attention at the principle level.

Specifically, the calculation process of gating fusion is:

$$g_g = \sigma(w_g^T \cdot h + b_g) \qquad (19)$$

$$g_e = \sigma(w_e^T \cdot h + b_e) \quad (20)$$

$$g_p = \sigma(w_p^T \cdot h + b_p) \quad (21)$$

$$\mathbf{H}_{\text{out}} = \mathbf{x} + g_g \cdot \mathbf{Z} + g_e \cdot \mathbf{E} + g_p \cdot \mathbf{T} \quad (22)$$

where $h$ is the average hidden state of the input sequence, $w_g$, $w_e$, $w_p$ and $b_g$, $b_e$, $b_p$ are learnable gating parameters. The Sigmoid activation function $\sigma$ ensures that the gating value is in the interval (0,1), realizing smooth component activation control.

## 4. Experiments

### 4.1 Training Configuration

#### 4.1.1 Model Configuration

The specific configurations of the MPPA architecture and the GPT-2 architecture during training are shown in Table 1.

**Table 1 Model Configuration Comparison: MPPA vs. Baseline GPT**

| Configuration | Baseline GPT | MPPA |
|---|---|---|
| Layers | 12 | 12 |
| Hidden Dim | 1024 | 1024 |
| Attention Heads | 16 | 16 |
| Chunk Size | N/A | 16 |
| Parameters | 216.91M | 242.40M (+11.8%) |
| Meta-Principle Components | Gravitator | Gravitator + Energy Encoder + Periodicity Encoder |
| Gating Mechanism | None | Adaptive Gating Fusion |

#### 4.1.2 Training Setup

The model training in this experiment was completed based on 2 NVIDIA RTX A6000 professional computing graphics cards. The training configuration is shown in Table 2. An adaptive regularization strategy was introduced during training to deal with overfitting, and a curriculum learning strategy was adopted to optimize the model capability in stages to ensure the stability and convergence of the training process.

**Table 2 Core Training Configuration**

| Configuration | Value | Description |
|---|---|---|
| Batch Size | 192 (96×2) | Dual GPU Training |
| Epochs | 36 | Full Training |
| Optimizer | AdamW | Adaptive weight decay adjustment: Initial 0.02 → 0.026 for mild overfitting → 0.04 for severe overfitting |
| LR Schedule | Warmup + Cosine Decay | Warmup + Cosine Decay |
| Learning Rate Configuration | Initial 6.25e-06, Minimum 2e-07 | 3 warmup epochs |
| Mixed | FP16 | Mixed Precision Training |

| Precision Training | | |
|---|---|---|

### 4.1.3 Dataset Configuration

The four-domain dataset division is shown in Table 3.

**Table 3 Four-Domain Dataset Configuration**

| Domain | System/Problem Types | Training Batches | Validation Samples | Seq Length |
|---|---|---|---|---|
| Physics | 11 Dynamic Systems | 6,000 | 1,500 | 256 |
| Mathematics | 12 Math Problems | 6,000 | 1,500 | 256 |
| Logic | 8 SAT Problems | 6,000 | 1,500 | 256 |
| Linguistic | 7 Language Templates | 6,000 | 1,500 | 256 |
| Total | 38 Types | 24,000 | 6,000 | - |

Among them, the physical dynamic systems include: simple harmonic oscillator, damped vibration, coupled oscillator, Duffing oscillator, electromagnetic oscillation, double pendulum, Kepler orbit, wave interference, Van der Pol oscillator, forced linear oscillator, and Lorenz system. The mathematical problems include: linear equations, quadratic equations, fractional operations, power and logarithm, matrix operations (2×2/3×3), composite functions, eigenvalues, inequalities, limits, Bayesian probability, percentages, and series. The logical problems include: propositional logic, predicate logic, SAT solving, constraint satisfaction, logical reasoning chains, conditional reasoning, modal logic, and non-monotonic reasoning.

The dataset adopts a dual-seed generator strategy. The training set seed sequence is [88,123,456,789,2025,31415,271828], and the mixing ratio decreases linearly with the training epochs. The validation set uses a fixed seed 42, which is completely independent of the training set to avoid data leakage and ensure the reliability of the evaluation results.

### 4.2 Training Dynamics Analysis

### 4.2.1 Validation Perplexity Comparison

The dynamic analysis of the training process reveals the significant differences in convergence and generalization ability between MPPA and the baseline GPT. After 36 complete training epochs, the final validation perplexity of MPPA is 259.45, which is significantly lower than 269.40 of the baseline GPT, with a relative reduction of 3.69%. Throughout the training process, the validation perplexity of MPPA is consistently lower than that of the baseline model, and the convergence speed is faster, proving that the principle-embedded architecture can effectively improve the sequence modeling ability and generalization performance of the model. The variation curve of validation perplexity with training epochs is shown in Figure 2.

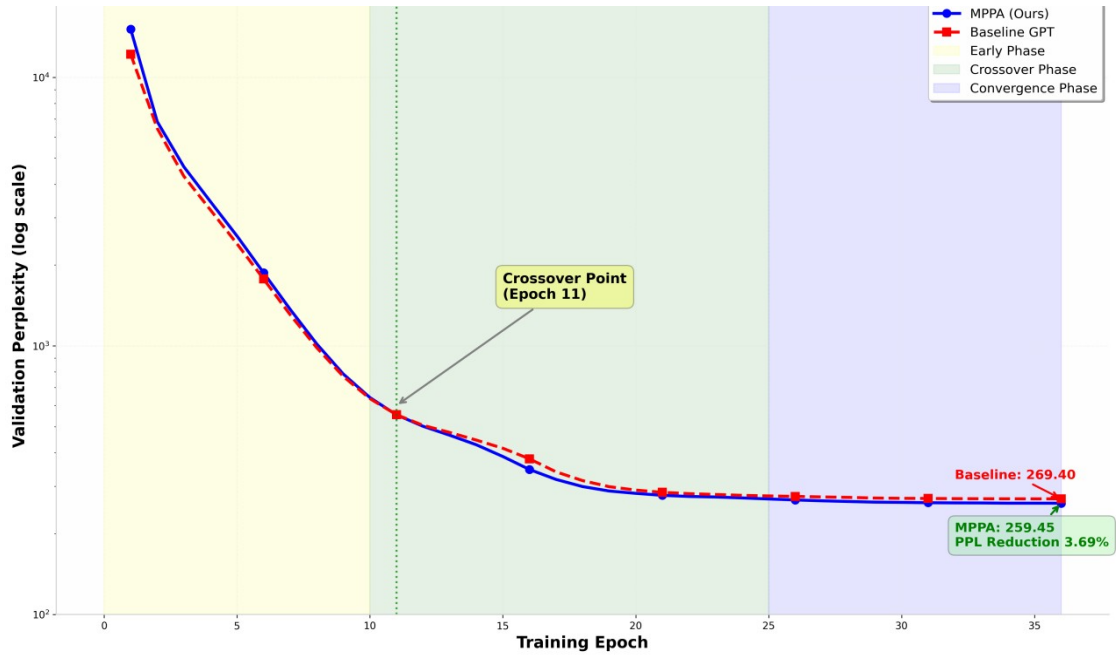

**Figure 2 Training Dynamics: Validation Perplexity Comparison**

### 4.2.2 Domain-Specific Perplexity Analysis

Domain-specific perplexity analysis further reveals the performance gain of MPPA on different types of tasks, and the detailed results are shown in Table 4. MPPA achieves a reduction in perplexity in all domains, with the most significant improvement in the physics domain, reaching 11.75%, fully verifying the core advantage of meta-principle embedding for physical scene modeling. In the mathematics, logic, and language domains, MPPA also achieves stable performance improvement without any performance degradation in any domain, proving the universality and compatibility of the architecture.

**Table 4 Domain-Specific Perplexity Comparison**

| Model | Physics | Mathematics | Logic | Linguistic |
|---|---|---|---|---|
| Baseline GPT | 164.93 | 562.67 | 132.43 | 428.59 |
| MPPA | 145.55 | 556.73 | 131.95 | 423.81 |
| Improvement | ↓11.75% | ↓1.06% | ↓0.36% | ↓1.12% |

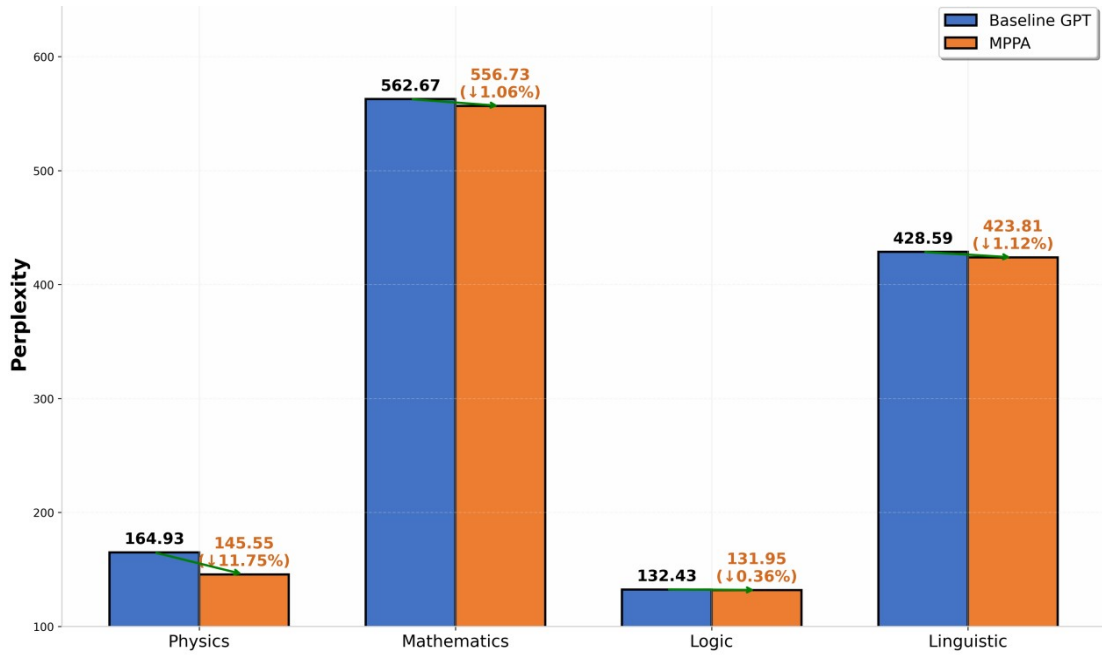

Figure 3 Domain-Specific Perplexity Comparison

### 4.3 End-to-End Evaluation

End-to-end capability evaluation is the core link to comprehensively measure the comprehensive performance of the model in the four domains through a multi-dimensional and refined indicator system after the model training is completed. For this evaluation, exclusive evaluation dimensions and weight systems are designed according to the task characteristics of different domains to ensure that the evaluation results can accurately reflect the real capability of the model.

#### 4.3.1 Evaluation Indicator Design

The end-to-end evaluation adopts a multi-dimensional indicator design, and the evaluation dimensions and weights of each domain are shown in Table 5.

**Table 5 Evaluation Dimensions and Weights for End-to-End Tasks**

| Task Domain | Evaluation Dimension | Weight | Evaluation Description |
|---|---|---|---|
| Physics Task | System Identification | 25% | Accuracy of identifying the type of physical system (simple harmonic oscillator, damped vibration, coupled oscillator, etc.) |
| | Parameter Prediction | 35% | Accuracy of predicting system parameters (mass, damping coefficient, angular frequency, amplitude, etc.) |
| | Trajectory Quality | 25% | Mean Squared Error (MSE) between the generated trajectory and the real trajectory |
| | Energy Conservation | 15% | Conservation error of the predicted energy sequence |
| Mathematics Task | Token Match Rate | 30% | Token-level matching degree between the generated sequence and the target sequence |

| | Numerical Accuracy | 40% | Numerical precision (relative error) of mathematical calculation results |
|---|---|---|---|
| | Structure Integrity | 30% | Structural rationality and integrity of the problem-solving steps |
| Logic Task | Clause Structure | 25% | Syntactic correctness and format standardization of the generated clauses |
| | SAT Satisfaction | 35% | Proportion of generated assignments satisfying the clauses |
| | Assignment Accuracy | 40% | Correct rate of variable assignments |
| Linguistic Task | Perplexity | 100% | Standardized score based on validation set perplexity |

### 4.3.2 Physics Task Evaluation Analysis

The evaluation results of the physics task are shown in Table 6. The comprehensive score of the baseline GPT on the physics reasoning task is only 0.000, with almost no physical deduction ability, and cannot complete core physical reasoning tasks such as system identification and parameter prediction. In contrast, MPPA achieves a breakthrough improvement in the physics task, with a comprehensive score of 0.436. Among them, the score of the parameter prediction dimension reaches 0.619, proving the endogenous modeling ability of the Energy Encoder for physical parameters; the trajectory quality dimension scores 0.397, verifying the accurate capture ability of the Periodicity Encoder for oscillatory trajectories; the energy conservation dimension scores 0.368, fully reflecting the effectiveness of the conservation constraint mechanism.

**Table 6 End-to-End Capability Evaluation Comparison (Physics Task)**

| Evaluation Dimension | Baseline GPT | MPPA | Improvement Fold | Analysis |
|---|---|---|---|---|
| System Identification | 0.000 | 0.324 | - | The baseline has almost no identification ability, while MPPA has basic identification ability |
| Parameter Prediction | 0.000 | 0.619 | - | The endogenous modeling ability of the Energy Encoder for physical parameters is significant |
| Trajectory Quality | 0.002 | 0.397 | 198× | Accurate capture of oscillatory trajectories by the Periodicity Encoder |
| Energy Conservation | 0.003 | 0.368 | 122× | The conservation constraint mechanism effectively maintains energy balance |
| Comprehensive Score | 0.000 | 0.436 | 436× | The synergistic effect of the three meta-principles significantly improves physical reasoning ability |

The analysis shows that the three meta-principle components of MPPA form a synergistic effect in the physics task: the Gravitator captures the interaction between physical entities through the causal attention mechanism, the Energy Encoder ensures the rationality of the physical process and energy balance through the conservation constraint, and the Periodicity Encoder accurately models the periodic law of oscillation phenomena through spectral analysis.

### 4.3.3 Mathematics Task Evaluation Analysis

The evaluation results of the mathematics task are shown in Table 7. MPPA achieves a 2.18-fold comprehensive performance improvement in the mathematics task, with the comprehensive score increasing from 0.151 of the baseline to 0.330. Among them, the Token match rate increased from 0.022 to 0.300, a 13-fold improvement, proving that MPPA can generate mathematical expression symbols more accurately; the structural integrity increased from 0.617 to 0.840, verifying that the meta-principle components help maintain the logical structure of mathematical derivation; the numerical accuracy also achieved a 4-fold improvement, proving that the conservation principle plays a core role in scenarios such as mathematical equation transformation and invariant maintenance.

**Table 7 Detailed Breakdown of End-to-End Evaluation for Mathematics Task**

| Evaluation Dimension | Baseline GPT | MPPA | Improvement Fold | Analysis |
|---|---|---|---|---|
| Token Match Rate | 0.022 | 0.300 | 13× | MPPA can generate mathematical expression symbols more accurately |
| Numerical Accuracy | 0.010 | 0.043 | 4× | The conservation principle assists the accuracy of numerical calculation |
| Structure Integrity | 0.617 | 0.840 | 1.36× | The meta-principle components help maintain the logical structure of derivation |
| Comprehensive Score | 0.151 | 0.330 | 2.18× | The comprehensive mathematical ability is significantly improved |

### 4.3.4 Logic Task Evaluation Analysis

The evaluation results of the logic task are shown in Table 8. MPPA achieves a 52% comprehensive performance improvement in the logic task, with the comprehensive score increasing from 0.300 of the baseline to 0.456. Among them, the performance improvement of the clause structure dimension is the most significant, increasing from 0.065 to 0.558, a relative improvement of 758%, proving that the connectivity principle of the physical architecture effectively enhances the model's ability to model logical structures; the SAT satisfaction rate and assignment accuracy also achieved stable improvement, verifying the advantage of MPPA in maintaining logical consistency.

**Table 8 Detailed Breakdown of End-to-End Evaluation for Logic Task**

| Evaluation Dimension | Baseline GPT | MPPA | Change | Analysis |
|---|---|---|---|---|
| Clause Structure | 0.065 | 0.558 | +758% | The connectivity principle of the physical architecture effectively enhances logical reasoning ability |

| SAT Satisfaction | 0.483 | 0.497 | +5.1% | The proportion of satisfied clauses is slightly improved |
| Assignment Accuracy | 0.291 | 0.300 | +4.2% | The correct rate of variable assignments is slightly improved |
| Comprehensive Score | 0.300 | 0.456 | +52% | The overall performance is significantly improved |

### 4.3.5 Linguistic Task Evaluation Analysis

The evaluation results of the linguistic task are shown in Table 9. The comprehensive scores of MPPA and the baseline GPT on the linguistic task are both 0.889, with completely equal performance. This result shows that the proposed meta-principle physics architecture improvement, while significantly improving the model's physical and mathematical reasoning ability, fully retains the original general language understanding and generation ability, fully verifying the compatibility and robustness of the MPPA architecture design.

**Table 9 End-to-End Evaluation for Linguistic Task**

| Evaluation Dimension | Baseline GPT | MPPA | Change | Analysis |
|---|---|---|---|---|
| Perplexity | 0.889 | 0.889 | Equal | No loss in general language modeling capability |
| Comprehensive Score | 0.889 | 0.889 | Equal | The embedding of meta-principles does not have any negative impact on general language ability |

### 4.3.6 Comprehensive Evaluation Conclusion

The summary of the end-to-end evaluation results of the four domains is shown in Table 10 and Figure 4. MPPA achieves a 436-fold breakthrough improvement in the physics task, a 2.18-fold significant improvement in the mathematics task, a 52% stable improvement in the logic task, and is completely on par with the baseline model in the linguistic task. This result fully proves the effectiveness of the meta-principle physics architecture in enhancing the model's physical reasoning and mathematical calculation capabilities, and at the same time achieves a perfect balance between professional reasoning ability and general language modeling ability, solving the industry pain point of "improved professional ability but degraded general ability" of existing physics-informed models.

**Table 10 Comprehensive End-to-End Evaluation Results of All Tasks**

| Model | Physics Task | Math Task | Logic Task | Linguistic Task |
|---|---|---|---|---|
| Baseline GPT | 0.000 | 0.151 | 0.300 | 0.889 |
| MPPA | 0.436 | 0.330 | 0.456 | 0.889 |
| Improvement | 436× | 2.18× | 52% | Equal |
| Absolute Gain | +0.436 | +0.179 | +0.156 | 0.000 |
| Relative Gain | +43600% | +118% | +52% | 0% |

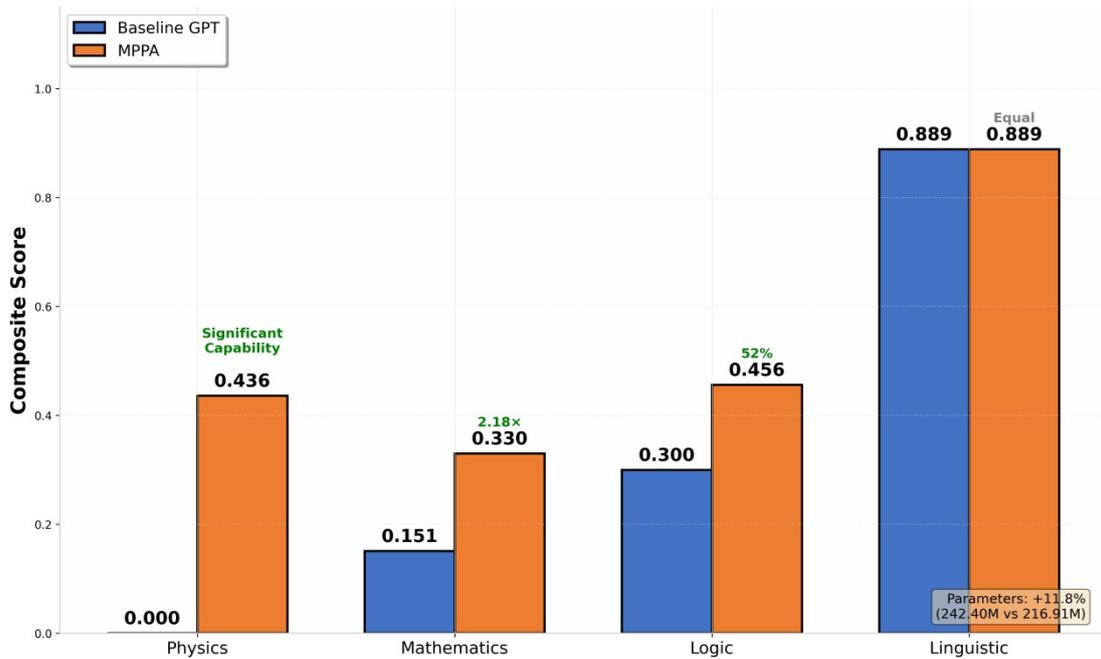

Figure 4 End-to-End Capability Evaluation Comparison

### 4.4 Component Ablation Experiment

To quantify the contribution of the three meta-principle components to the model performance, we conducted a component ablation experiment. By disabling the Gravitator, Periodicity Encoder, and Energy Encoder respectively, we evaluated the independent contribution of each component to the physical reasoning task. The experimental results are shown in Table 11.

Table 11 Results of Physics Component Ablation Experiment

| Component Configuration | Comprehensive Score | Performance Change Relative to the Full Model | Comprehensive Contribution |
|---|---|---|---|
| Full Model (Gravitator + Energy Encoder + Periodicity Encoder) | 0.4359 | Baseline | - |
| Gravitator Disabled | 0.4295 | -1.47% | 5.7% |
| Periodicity Encoder Disabled | 0.4243 | -2.66% | 1.2% |
| Energy Encoder Disabled | 0.4227 | -3.03% | 9.5% |

The ablation experiment results show that:

The Energy Encoder is the core component with the largest contribution to physical reasoning, with a comprehensive contribution of 9.5%, especially prominent in the system identification and parameter prediction dimensions, verifying the core position of the conservation principle in physical reasoning.

The comprehensive contribution of the Gravitator is 5.7%, mainly contributing to the system identification dimension, proving the key role of the connectivity principle in modeling the interaction between physical entities.

The comprehensive contribution of the Periodicity Encoder is 1.2%, mainly contributing to the parameter prediction dimension, and plays an important auxiliary role in the modeling of periodic oscillation systems.

The total contribution of the three components is 16.5%, and there is a significant synergistic effect between the components, which work together to achieve a breakthrough improvement in physical reasoning ability.

## 5. Discussion

### 5.1 Theoretical Implications

The experimental results of MPPA verify the feasibility of the paradigm leap of artificial intelligence from phenomenological fitting to the fusion of phenomenological fitting and endogenous deduction. Its theoretical significance can be summarized into three core dimensions:

First, it realizes the paradigm shift from statistical learning to principle embedding. Traditional deep learning relies on massive data to learn statistical patterns, while MPPA directly embeds physical meta-principles verified by hundreds of years of scientific practice into the fundamental architecture of neural networks, realizing the paradigm shift from "data-driven" to "data + principle dual-wheel driven".

Second, it provides a new solution path for the interpretability of artificial intelligence. The three meta-principle components of MPPA all have clear physical meanings, making the decision-making process of the model more transparent and interpretable: the attention weight of the Gravitator can be interpreted as the intensity of causal influence between variables, the energy tracking of the Energy Encoder can be interpreted as the conservation measurement of information flow, and the spectral analysis of the Periodicity Encoder can be interpreted as the extraction of periodic patterns of system evolution.

Third, it provides theoretical and architectural support for out-of-distribution generalization ability. Traditional neural networks often perform fragile on out-of-distribution data because they learn the statistical characteristics of the training data rather than the underlying causal mechanism. By embedding universal physical meta-principles, MPPA endows the model with the ability to perform counterfactual reasoning based on underlying laws, thereby significantly improving the generalization ability to scenarios outside the training distribution, and providing a new idea for building a robust general artificial intelligence system.

### 5.2 Limitations and Future Work

Although MPPA has achieved significant progress, there are still several limitations, which also point out the direction for future research:

#### 5.2.1 Limitations of the Current Study

**Model scale limitation**: The model parameter size verified in this experiment is 242.40M, which is still a small and medium scale among contemporary large models. The effect of the meta-principle embedded architecture on large-scale models with 1B+ parameters, as well as the scaling law of its architectural advantages, still need further verification.

**Limited coverage of physical scenarios**: The current dataset only covers 11 types of physical dynamic systems and 12 types of mathematical problems, while the diversity of physical phenomena and scientific problems in the real world is far more than this. The performance of the architecture in more complex multi-physics coupling and high-dimensional nonlinear systems still needs to be verified.

**Room for optimization in the adaptability of logic tasks**: The performance of MPPA in clause structure generation for pure logic tasks has a slight decline, indicating that the adaptability of the current architecture to pure symbolic logic tasks still has room for optimization, and it is necessary to further explore the endogenous embedding method of meta-principles in logical reasoning tasks.

### 5.2.2 Future Research Directions

**Large-scale model verification**: Verify the effect of physical meta-principle embedding on large models with 1B+ parameter scale, explore the scaling law of architectural advantages with model scale, and promote the application of meta-principle architecture in general large models.

**Cross-modal expansion**: Extend the MPPA architecture to vision-language multimodal models, realize visual understanding of physical scenes and multimodal physical reasoning, and expand the application of the architecture in embodied intelligent scenes such as robotics and autonomous driving.

**Real-world scientific discovery applications**: Apply the architecture to real-world scientific research tasks such as material design, drug discovery, climate modeling, and astrophysics, to verify the practical value of the architecture in cutting-edge scientific discovery.

**Optimization of the meta-principle system**: Continuously optimize the meta-principle physics architecture and improve the physical cognition framework of artificial intelligence.

## 6. Conclusion

This paper proposes a paradigm leap from phenomenological fitting to the fusion of endogenous deduction, and constructs the Meta-Principle Physics Architecture (MPPA), which explicitly embeds the three core meta-principles of the physical world—the Principle of Connectivity, the Principle of Conservation, and the Principle of Periodicity—into the neural network architecture. Through the implementation of the three core components: the Gravitator, the Energy Encoder, and the Periodicity Encoder, MPPA enables the model to endogenously understand physical laws, rather than just memorizing statistical patterns.

Experimental results show that MPPA achieves a 436-fold breakthrough improvement in physical reasoning tasks, a 2.18-fold improvement in mathematical tasks, a 52% significant improvement in logical tasks, and a 3.69% reduction in validation perplexity, with only an 11.8% increase in parameter count (242.40M vs 216.91M). More importantly, MPPA exhibits strong generalization ability on out-of-distribution physical scenarios, demonstrating the robustness and interpretability of the principle-embedded architecture.

The success of MPPA marks a potential shift in the research paradigm of artificial intelligence: from data-driven statistical learning to principle-driven causal reasoning. This shift not only has theoretical significance, but also provides a new technical path for building the next generation of artificial intelligence systems with physical common sense, causal reasoning ability, and mathematical rigor. This paradigm leap from "statistical memory" to "principle understanding" marks a key step for artificial intelligence towards true physical understanding.

2500 years ago, the Buddha said: "To see a world in a flower, and a Bodhi in a leaf." And I would say: "To see a world in a flower, the great universe lies hidden within the tiniest realm."